# Max-Pooling Dropout for Regularization of Convolutional Neural Networks

Haibing Wu and Xiaodong Gu

Department of Electronic Engineering, Fudan University, Shanghai 200433, China
`haibingwu13@fudan.edu.cn, xdgu@fudan.edu.cn`

**Abstract.** Recently, dropout has seen increasing use in deep learning. For deep convolutional neural networks, dropout is known to work well in fully-connected layers. However, its effect in pooling layers is still not clear. This paper demonstrates that *max-pooling dropout* is equivalent to randomly picking activation based on a multinomial distribution at training time. In light of this insight, we advocate employing our proposed probabilistic weighted pooling, instead of commonly used max-pooling, to act as model averaging at test time. Empirical evidence validates the superiority of probabilistic weighted pooling. We also compare max-pooling dropout and stochastic pooling, both of which introduce stochasticity based on multinomial distributions at pooling stage.

## 1 Introduction

Deep convolutional neural networks (CNNs) have recently been substantially improving on the state of art in computer vision. A standard CNN consists of alternating convolutional and pooling layers, with fully-connected layers on top. Compared to regular feed-forward networks with similarly-sized layers, CNNs have much fewer connections and parameters due to the local-connectivity and shared-filter architecture in convolutional layers, so they are far less prone to over-fitting. Another nice property of CNNs is that pooling operation provides a form of translation invariance and thus benefits generalization. Despite these attractive qualities and despite the fact that CNNs are much easier to train than other regular, deep, feed-forward neural networks, big CNNs with millions or billions of parameters still easily overfit relatively small training data.

Dropout [1] is a recently proposed regularizer to fight against over-fitting. It is a regularization method that stochastically sets to zero the activations of hidden units for each training case at training time. This breaks up co-adaptions of feature detectors since the dropped-out units cannot influence other retained units. Another way to interpret dropout is that it yields a very efficient form of model averaging where the number of trained models is exponential in that of units, and these models share the same parameters. Dropout has also inspired other stochastic model averaging methods such as stochastic pooling [4], drop-connect [5] and maxout networks [3].

Although dropout is known to work well in fully-connected layers of convolutional neural nets [1, 5, 6], its effect in pooling layers is, however, not well studied. This paper shows that using max-pooling dropout at training time is equivalent to sampling activation based on a multinomial distribution, and the distribution has a tunable parameter $p$ (the retaining probability). In light of this, probabilistic weighted pooling is proposed and employed at test time to efficiently average all possibly max-pooling dropout trained networks. Our empirical evidence confirms the superiority of probabilistic weighted pooling over max-pooling. Like fully-connected dropout, the number of possible max-pooling dropout models also grows exponentially with the increase of the number of hidden units that are fed into pooling layers, but decreases with the increase of pooling region's size.

As both stochastic pooling [4] and max-pooling dropout randomly sample activation based on multinomial distributions at pooling stage, it becomes interesting to compare their performance. Experimental results show that stochastic pooling performs between max-pooling dropout with different retaining probabilities, yet max-pooling dropout with typical retaining probabilities often outperforms stochastic pooling by large margins.

In this paper, dropout on the input to max-pooling layers is also called *max-pooling dropout* for brevity. Similarly, dropout on the input to fully-connected layers is called *fully-connected dropout*.

## 2    Related Work

Dropout is a new regularization technique that has been more recently employed in deep learning. Pioneering work by Hinton et al. [1] only applied dropout to fully connected layers. It was the reason they provided that the convolutional shared-filter architecture was a drastic reduction in the number of parameters and thus reduced its possibility to overfit in convolutional layers. Krizhevsky et al. [6] trained a very big convolutional neural net to classify 1.2 million ImageNet images. Two primary methods were used to reduce over-fitting in their experiments. The first one was data augmentation, an easiest and most commonly used approach to reduce over-fitting on image data. Dropout was exactly the second one. Also, it was only used in fully-connected layers.

Stochastic pooling [4] is a dropout-inspired regularization method. Instead of always capturing the strongest activity within each pooling region as max-pooling does, stochastic pooling randomly picks the activations according to a multinomial distribution.

Maxout network [3] is another model inspired by dropout. Combining with dropout, maxout networks have been shown to achieve best results on five benchmark datasets. However, the authors did not train maxout networks without dropout. Besides, they did not train the rectified counterparts with dropout and directly compare it with maxout networks. Dropout has also motivated other stochastic model averaging methods, such as drop-connect [5] and adaptive dropout [8].

## 3    Max-Pooling Dropout

We now demonstrate that max-pooling dropout is equivalent to sampling activation according to a multinomial distribution at training time. Basing on this interpretation, we propose to use probabilistic weighted pooling at test time.

### 3.1 Max-Pooling Dropout at Training Time

Consider a standard CNN composed of alternating convolutional and pooling layers, with fully-connected layers on top. On each presentation of a training example, if layer $l$ is followed by a pooling layer, the forward propagation without dropout can be described as

$$a_j^{(l+1)} = Pool(a_1^{(l)},..., a_i^{(l)},..., a_n^{(l)}), i \in R_j^{(l)}. \quad (1)$$

Here $R_j^{(l)}$ is pooling region $j$ at layer $l$ and is the activity of each neuron within it. $n = |R_j^{(l)}|$ is the number of units in $R_j^{(l)}$. $Pool()$ denotes the pooling function. Pooling operation provides a form of spatial transformation invariance as well as reduces the computational complexity for upper layers. An ideal pooling method is expected to preserve task-related information while discarding irrelevant image details. Two popular choices are average- and max-pooling. Average-pooling takes all activations in a pooling region into consideration with equal contributions. This may downplay high activations as many low activations are averagely included. Max-pooling only captures the strongest activation, and disregards all other units in the pooling region. We now show that employing dropout in max-pooling layers avoids both disadvantages by introducing stochasticity.

With dropout, the forward propagation becomes

$$\hat{a}^{(l)} \sim m^{(l)} * a^{(l)}, \quad (2)$$

$$a_j^{(l+1)} = Pool(\hat{a}_1^{(l)},..., \hat{a}_i^{(l)},..., \hat{a}_n^{(l)}), i \in R_j^{(l)}. \quad (3)$$

Here $*$ denotes element wise product and $m^{(l)}$ is a binary mask with each element $m_i^{(l)}$ drawn independently from a Bernoulli distribution. This mask is multiplied with activations $a^{(l)}$ in a pooling region at layer $l$ to produce dropout-modified activations $\hat{a}^{(l)}$. The modified activations are then passed to pooling layers. Fig. 1 presents a concrete example to illustrate the effect of dropout in max-pooling layers. Clearly, without dropout, the strongest activation in a pooling regions is always selected as the pooled activation. With dropout, it is not necessary that the strongest activation being the output. Therefore, max-pooling at training time becomes a stochastic procedure. To formulate such stochasticity, suppose the activations $(a_1^{(l)}, a_2^{(l)},..., a_n^{(l)})$ in each pooling region $j$ are reordered in non-decreasing order, i.e., $0 \leq a_1'^{(l)} \leq a_2'^{(l)} ... \leq a_n'^{(l)}$. With dropout, each unit in the pooling region could be possibly set to zero with probability of $q = 1 - p$ is the dropout probability, and $p$ is the retaining probability). As a result, $a_i'^{(l)}$ will be selected as the pooled activation on condition that (1) $a_{i+1}'^{(l)}, a_{i+2}'^{(l)},..., a_n'^{(l)}$ are dropped out, and (2) $a_i'^{(l)}$ is retained. This event occurs with probability of $p_i$ according to probability theory:

$$\Pr(a_j^{(l+1)} = a_i'^{(l)}) = p_i = pq^{n-i}, (i=1,2,...,n). \quad (4)$$

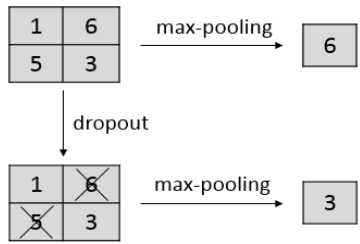

**Fig. 1.** An illustrating example showing the procedure of max-pooling dropout. The activation in the pooling region is 1, 6, 5 and 3 respectively. Without dropout, the strongest activation 6 is always selected as the output. With dropout, each unit in the pooling region could be possibly dropped out. In this example, only 1 and 3 are retained, then 3 will be the pooled output.

A special event occurring with probability of $p_0 = q^n$ is that all the units in a pooling region is dropped out, and the pooled output becomes $a_0'^{(l)} = 0$. Therefore, performing max-pooling over the dropout-modified pooling region is exactly sampling from the following multinomial distribution to select an index $i$, then the pooled activation is simply $a_i'^{(l)}$:

$$a_j^{(l+1)} = a_i'^{(l)}, \text{ where } i \sim Multinomial\ (p_0, p_1, p_2,..., p_n). \quad (5)$$

Let $s$ be the size of a feature map at layer $l$ (with $r$ feature maps), and $t$ be the size of pooling regions. The number of pooling region is therefore $rs/t$ for non-overlapping pooling. Each pooling region provides $t+1$ choices of the indices, then the number of possibly trained models $C$ at layer $l$ is

$$C = (1+t)^{rs/t} = \left(\sqrt[t]{1+t}\right)^{rs} = b(t)^{rs}. \quad (6)$$

So the number of possibly max-pooling dropout trained models is exponential in the number of units that are fed pooling max-pooling layers, and the base $b(t)$ $(1 < b(t) = \sqrt[t]{1+t}) \leq 2)$ depends on the size of pooling regions. Obviously, with the increase of the size of pooling regions, the base $b(t)$ decreases, and the number of possibly trained models becomes smaller. Note that the number of possibly fully-connected dropout trained models is also exponential in the number of units that are fed into fully-connected layers, but with 2 as the base.

### 3.2 Probabilistic Weighted Pooling at Test Time

Using dropout in fully-connected layers during training, the whole network containing all the hidden units should be used at test time, but with their outgoing weights halved to compensate for the fact that twice as many of them are active [1], or with their activations halved. Using max-pooling dropout during training, one might intuitively pick as output the strongest activation multiplied by the retaining probability:

$$a_j^{(l+1)} = p \times \max(a_1^{(l)},..., a_i^{(l)},..., a_n^{(l)}). \quad (7)$$

Since the strongest activation in a pooling region is scaled down by the retaining probability, we call this *scaled max-pooling*.

At test time, scaled max-pooling generally works well in practice, but is not the optimal. Instead we propose to use probabilistic weighted pooling to efficiently get a more accurate approximation of averaging all possibly trained dropout networks. In this pooling scheme, the pooled activity is linear weighted summation over activations in each region:

$$a_j^{(l+1)} = \sum_{i=0}^{n} p_i a_i'^{(l)} = \sum_{i=1}^{n} p_i a_i'^{(l)}. \tag{8}$$

Here $p_i$ is exactly the probability calculated by Eqn. (4). This type of probabilistic weighted summation can be interpreted as an efficient form of model averaging where each selection of index $i$ corresponds to a different model. Empirical evidence will confirm that probabilistic weighted pooling is a more accurate approximation of averaging all possible dropout models than scaled max-pooling.

## 4 Empirical Evaluations

Experiments are conducted on three datasets: MNIST, CIFAR-10 and CIFAR-100. MNIST consists of 28x28 pixel grayscale images, each containing a digit 0 to 9. There are 60,000 training and 10,000 test examples. We do not perform any preprocessing except scaling the pixel values to [0, 1]. The CIFAR-10 dataset [2] consists of ten classes of natural images with 50,000 examples for training and 10,000 for testing. Each example is a 32x32 RGB image taken from the tiny images dataset collected from the web. CIFAR-100 is just like CIFAR-10, but with 100 categories. We also scale to [0, 1] for CIFAR-10 and CIFAR-100 and subtract the mean value of each channel computed over the dataset for each image.

We use rectified linear function [7] for convolutional and fully-connected layers, and softmax activation function for the output layer. More commonly used sigmoidal and tanh nonlinearities are not adopted due to gradient vanishing problem with them. Our models are trained using stochastic mini-batch gradient descent with a batch size of 100, momentum of 0.95, learning rate of 0.1 to minimize the cross entropy loss. The weights in all layers are initialized from a zero-mean Gaussian distribution with 0.1 as standard deviation and the constant 0 as the neuron biases in all layers.

The CNN architecture for MNIST is 1x28x28-20C5-2P2-40C5-2P2-1000N-10N, which represents a CNN with 1 input image of size 28x28, a convolutional layer with 20 feature maps and 5x5 filters, a pooling layer with pooling region 2x2 and stride 2, a convolutional layer with 40 feature maps and 5x5 filters, a pooling layer with pooling region 2x2 and stride 2, a fully-connected layer with 1000 hidden units, and an output layer with 10 units (one per class). The architecture for CIFAR-10 is 3x32x32-96C5-3P2-128C3-3P2-256C3-3P2-2000N-2000N-10N. The architecture for CIFAR-100 is the same with CIFAR-10 except with 100 output units.

### 4.1 Probabilistic Weighted Pooling vs. (Scaled) Max-Pooling

We initially validate the superiority of probabilistic weighted pooling over max-pooling and scaled max-pooling using MNIST. The CNNs are trained for 1000 epochs. For max-pooling dropout, CNN models are trained with different retaining probabilities. Fig. 2 compares the test performances produced by different pooling methods at test time. Generally, probabilistic weighted pooling performs better than max-pooling and scaled max-pooling with different retaining probabilities. For small $p$ (the retaining probability), max-pooling and scaled max-pooling performs very poorly; probabilistic weighted pooling is considerably better. With the increase of $p$, the performance gap becomes smaller. This is not surprising as the pooled outputs for different pooling methods are close to each other for large $p$. An extreme case is that when $p = 1$, scaled max-pooling and probabilistic weighted pooling are exactly the same with max-pooling.

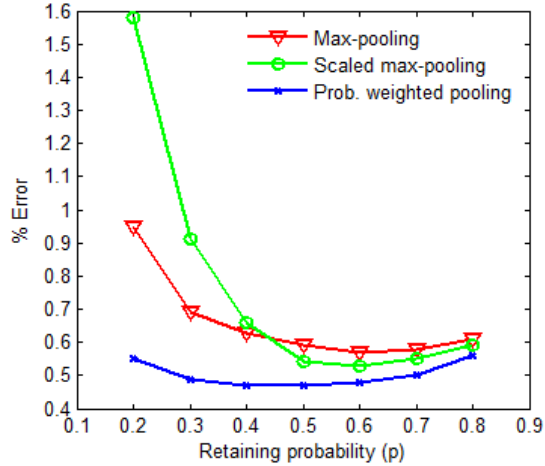

**Fig. 2.** MNIST test errors for different pooling methods at test time. Max-pooling dropout is used to train CNN models with different retaining probabilities at training time.

We then compares different pooling methods at test time for max-pooling dropout trained models on CIFAR-10 and CIFAR-100. The retaining probability is set to 0.3, 0.5 and 0.7 respectively. At test time, max-pooling, scaled max-pooling and probabil-istic weighted pooling are respectively used to act as model averaging. Fig. 3 presents the test performance of these pooling methods. Again, for small retaining probability $p = 0.3$, scaled max-pooling and probabilistic weighted pooling perform poorly. Probabilistic weighted pooling is the best performer with different retaining probabilities. The increase of $p$ narrows different pooling methods' performance gap.

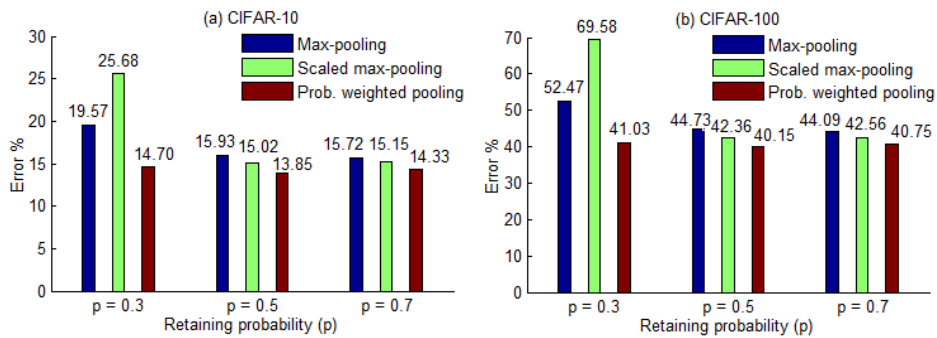

**Fig. 3.** CIFAR-10 and CIFAR-100 test errors for different pooling methods at test time. Max-pooling dropout is used to train CNNs with different retaining probabilities at training time.

### 4.2    Max-Pooling Dropout vs. Stochastic Pooling

Similar to max-pooling dropout, stochastic pooling [4] also randomly picks activation according to a multinomial distribution at training time, and also involves probabilistic weighting at test time. More concretely, at training time it first computes the probability $p_i$ for each unit within pooling region $j$ at layer $l$ by normalizing the activations:

$$p_i = \frac{a_i^{(l)}}{\sum_{k=1}^{n} a_k^{(l)}}, \ (i = 1,2,...,n). \tag{9}$$

It then samples from a multinomial distribution based on $p_i$ to select an index $i$ in the pooling region. The pooled activation is simply $a_i^{(l)}$:

$$a_j^{(l+1)} = a_i^{(l)}, \text{ where } i \sim Multinomial\ (p_1, p_2,..., p_n). \tag{10}$$

At test time, probabilistic weighting is adopted to act as model averaging. That is, the activations in each pooling region are weighted by the probability $p_i$ and summed:

$$a_j^{(l+1)} = \sum_{i=1}^{n} p_i a_i^{(l)}. \tag{11}$$

One may have found that stochastic pooling bears much resemblance to max-pooling dropout, as both involve stochasticity at pooling stage. We are therefore very interested in their performance differences. To compare their performances, we train CNN models with different retaining probabilities on MNIST, CIFAR-10 and CIFAR-100. For max-pooling dropout trained models, only probabilistic weighted pooling is used at test time. Fig. 4 compares the test performances of max-pooling dropout with different retaining probabilities against stochastic pooling. The relation between the performance of max-pooling dropout and the retaining probability $p$ is a U-shape. If $p$ is too small or too large, max-pooling dropout performs poorer than stochastic pooling. Yet max-pooling dropout with typical $p$ (around 0.5) outperforms stochastic pooling by a large margin. Therefore, although stochastic pooling is hyper-parameter free and this saves the tuning of retaining probability, its performance is often inferior to max-pooling dropout.

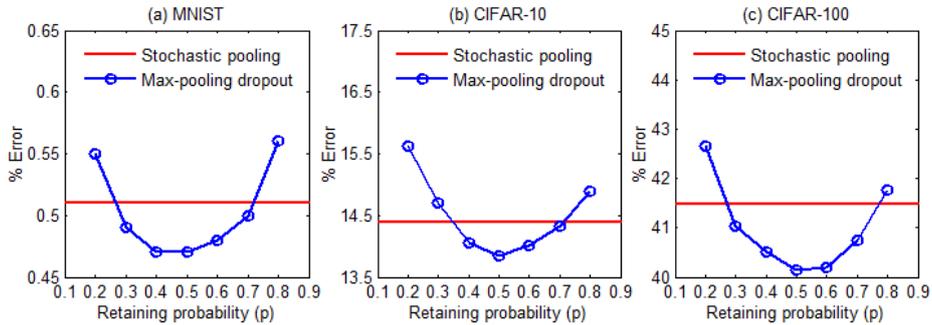

**Fig. 4.** MNIST, CIFAR-10 and CIFAR-100 test errors for max-pooling dropout with different retaining probabilities against stochastic pooling.

## 5      Conclusions

This paper mainly addresses the problem of understanding and using dropout on the input to max-pooling layers of convolutional neural nets. At training time, max-pooling dropout is equivalent to randomly picking activation according to a multinomial distribution, and the number of possibly trained networks is exponential in the number of input units to the pooling layers. At test time, a new pooling method, probabilistic weighted pooling, is proposed to act as model averaging. Experimental evidence confirms the benefits of using max-pooling dropout, and validates the superiority of probabilistic weighted pooling over max-pooling and scaled max-pooling. Considering that stochastic pooling is similar to max-pooling dropout, we empirically compare them and show that the performance of stochastic pooling is between those produced by max-pooling dropout with different retaining probabilities.

## Acknowledgements

This work was supported in part by National Natural Science Foundation of China under grant 61371148 and Shanghai National Natural Science Foundation under grant 12ZR1402500.